%% file: main.tex
\documentclass[sigconf]{acmart}

\def\BibTeX{{\rm B\kern-.05em{\sc i\kern-.025em b}\kern-.08emT\kern-.1667em\lower.7ex\hbox{E}\kern-.125emX}}
\usepackage{caption}
\usepackage{subcaption}    
\usepackage{caption}
\usepackage{flafter} 
\usepackage{float}
\usepackage{lipsum}
\usepackage{multirow}
\usepackage{pifont}
\usepackage{rotating}

\copyrightyear{2024}
\acmYear{2024}
\setcopyright{rightsretained}
\acmConference[NSysS '24]{11th International Conference on Networking, Systems, and Security}{December 19--21, 2024}{Khulna, Bangladesh}
\acmBooktitle{11th International Conference on Networking, Systems, and Security (NSysS '24), December 19--21, 2024, Khulna, Bangladesh}
\acmPrice{}
\acmDOI{10.1145/3704522.3704555}
\acmISBN{979-8-4007-1158-9/24/12}

\begin{document}

\title{Short Paper: Revealing the Self: Brainwave-Based Human Trait Identification}

\author{Md Mirajul Islam}
\affiliation{%
  \institution{Bangladesh University of Engineering and Technology}
  \streetaddress{}
  \city{Dhaka}
  \country{Bangladesh}
  \state{}
  \postcode{}
}
\email{mirajul1995@gmail.com}
\authornotemark[1]

\author{Md Nahiyan Uddin}
\affiliation{%
  \institution{Bangladesh University of Engineering and Technology}
  \streetaddress{}
  \city{Dhaka}
  \country{Bangladesh}
  \state{}
  \postcode{}
}
\email{nahiyan.uddin.102@gmail.com}
\authornote{Both authors contributed equally to this work.}

\author{Maoyejatun Hasana}
\affiliation{%
  \institution{Bangladesh University of Engineering and Technology}
  \streetaddress{}
  \city{Dhaka}
  \country{Bangladesh}
  \state{}
  \postcode{}
}
\email{hasana004@gmail.com}

\author{Debojit Pandit}
\affiliation{%
  \institution{Bangladesh University of Engineering and Technology}
  \streetaddress{}
  \city{Dhaka}
  \country{Bangladesh}
  \state{}
  \postcode{}
}
\email{debojitpanditdip@gmail.com}

\author{Nafis Mahmud Rahman}
\affiliation{%
  \institution{Sunbeams}
  \streetaddress{}
  \city{Dhaka}
  \country{Bangladesh}
  \state{}
  \postcode{}
}
\email{nafis.mahmud.rahman@gmail.com}

\author{Sriram Chellappan}
\affiliation{%
  \institution{University of South Florida}
  \streetaddress{}
  \city{Tampa, FL}
  \country{USA}}
\email{sriramc@usf.edu}

\author{Sami Azam}
\affiliation{%
  \institution{Charles Darwin University}
  \streetaddress{}
  \city{Darwin}
  \country{Australia}}
\email{sami.azam@cdu.edu.au}

\author{A. B. M. Alim Al Islam}
\affiliation{%
  \institution{Bangladesh University of Engineering and Technology}
  \streetaddress{}
  \city{Dhaka}
  \country{Bangladesh}
  \state{}
  \postcode{}
}
\email{alim_razi@cse.buet.ac.bd}

\renewcommand{\shortauthors}{Mirajul and Nahiyan, et al.}

\begin{abstract}
People exhibit unique emotional responses. In the same scenario, the emotional reactions of two individuals can be either similar or vastly different. For instance, consider one person's reaction to an invitation to smoke versus another person's response to a query about their sleep quality. The identification of these individual traits through the observation of common physical parameters opens the door to a wide range of applications, including psychological analysis, criminology, disease prediction, addiction control, and more. While there has been previous research in the fields of psychometrics, inertial sensors, computer vision, and audio analysis, this paper introduces a novel technique for identifying human traits in real time using brainwave data. To achieve this, we begin with an extensive study of brainwave data collected from $80$ participants using a portable EEG headset. We also conduct a statistical analysis of the collected data utilizing box plots. Our analysis uncovers several new insights, leading us to a groundbreaking unified approach for identifying diverse human traits by leveraging machine learning techniques on EEG data. Our analysis demonstrates that this proposed solution achieves high accuracy. Moreover, we explore two deep-learning models to compare the performance of our solution. Consequently, we have developed an integrated, real-time trait identification solution using EEG data, based on the insights from our analysis. To validate our approach, we conducted a rigorous user evaluation with an additional $20$ participants. The outcomes of this evaluation illustrate both high accuracy and favorable user ratings, emphasizing the robust potential of our proposed method to serve as a versatile solution for human trait identification.

\end{abstract}

\ccsdesc[500]{Computer systems organization~Embedded systems}
\ccsdesc[300]{Computer systems organization~Redundancy}
\ccsdesc{Computer systems organization~Robotics}
\ccsdesc[100]{Networks~Network reliability}

\keywords{Brainwaves, EEG, machine learning, deep learning, real-time, behavior.}

\maketitle

\section{Introduction}
Human traits reflect on their behavioral and emotional patterns that evolve from biological and environmental factors. Naturally, the traits of every individual are unique to that person and influenced by a diverse set of factors, including genetics, education, experiences, age, climate, and more. Recently, the issue of mechanisms for identifying individual traits has gained significant attention across various fields, such as job suitability ~\cite{cole2008}, law enforcement ~\cite{campbell2004}, disease detection ~\cite{joseph2016}, psychological counseling ~\cite{guo2008}, and many others.

There are a few human traits or activities, such as smoking or alcohol consumption, that can be identified through some physical diagnostic tests such as blood tests ~\cite{gnann2009}, drug tests ~\cite{Connor871}, and the like. Nevertheless, these methods are typically very expensive ~\cite{ash2001} and specifically designed for identifying a particular trait or activity ~\cite{rog2005}. Alternative approaches involve questionnaires or interviews, which offer more cost-effective means of identifying human traits ~\cite{muba2011}. However, these approaches are also trait-specific and often demand experts to analyze the responses to the questions ~\cite{mir2013}. Consequently, the existing approaches fall short of providing a unified solution that can be applied comprehensively. Thus, the challenge of identifying diverse human traits in a uniform and ubiquitous manner remains unsolved in the literature. 

To this extent, in this paper, we address the issue by introducing the exploitation of the notion of brainwave analytics for identifying human traits. We specifically focus on brainwave analytics, as brainwaves are known to be molded as per human traits and activities ~\cite{deary2010}. Recognizing this, we employ various machine-learning techniques to perform the identification of human traits based on brainwave data. Our analysis is conducted using EEG data collected from 80 individuals through a portable EEG headset. Our approach involves the application of several established machine learning techniques with a primary focus on the identification of their distinct traits. Given that humans express a wide range of emotions, which can also impact brainwave data ~\cite{uchida2015}, we intentionally induce various emotions in the subjects during EEG data collection. 
We observe that machine learning-based classifications applied to the collected EEG data demonstrate significant accuracy in trait identification. One of the most appealing aspects of this approach is its capacity to simultaneously identify various human traits. This is possible because each human trait potentially exerts a unique influence on the brainwave signals recorded by the EEG headset, which can be discerned by machine learning techniques. Intuitively, the human brain should exhibit a close connection with human traits and activities, and the same holds true for brainwaves. Brainwaves can be classified into eight categories based on their frequencies ~\cite{silva1991}. These wave signals can be measured through electroencephalogram (EEG) ~\cite{teplan2002}. It remains challenging to determine which trait(s) may impact specific signals and to what extent. Therefore, we consider all of the signals together in the process of identifying human traits. In this study, we explore a total of 14 different human traits including those influenced by heredity. These traits are religious practice, smoking, religious beliefs, physical exercise, family history of diabetes, family history of heart disease, family history of brain stroke, fast food consumption, high-fat levels, high sugar intake, outdoor game participation, sleep-related issues, sleep patterns, and vegetable consumption. We collected this data from 80 different individuals and recorded their brainwave signals in various emotional states. 

Subsequently, we trained a range of machine-learning models and two deep-learning models using the collected brainwave data. Our analysis consistently demonstrates significant accuracy in trait identification. In turn, we developed a comprehensive, integrated solution for real-time human trait detection. We conducted experiments with an additional 20 different individuals using this solution, which resulted in substantial accuracy in identifying their traits. Moreover, our solution's user-friendly interface garnered positive ratings from the study participants. The combined strengths of accuracy, user satisfaction, and the lightweight design of our proposed solution make it suitable for widespread deployment in trait identification applications. 

Based on our research, we make the following contributions in this paper:

\begin{itemize}
    \item We collect brainwave signals from 80 individuals using an EEG headset while inducing four distinct emotional states. To achieve this, we create a customized survey application to automate the entire data collection and trait identification process. We seamlessly integrate the EEG headset with the application. 
    \item We perform statistical analysis on the collected data utilizing box plots, thereby providing valuable insights into the distribution and variability of EEG signals across different emotional conditions.
    \item Subsequently, we employ Auto-WEKA ~\cite{autoweka2017}, a data mining software, to identify the most effective machine learning technique for the task of trait identification. We also perform necessary hyperparameter tuning during the analysis.
    \item We compare the performance of the Auto-WEKA models against two deep-learning models - Long Short-Term Memory (LSTM) and Bidirectional Long Short-Term Memory (BiLSTM). 
    \item Finally, we develop a real-time unified trait identification application. The user evaluation demonstrates substantial accuracy and favorable user ratings, confirming the effectiveness of our proposed technique in real-time trait identification.
\end{itemize}

\section{Background and Motivation}
Our research builds on prior examinations of human brain activity to indicate its relationship with human emotion and physical activities. Neurons in the human brain function by communicating with each other through electrical impulses. When these neurons become active, they generate local electrical currents. This activity, driven collectively by the flow of electrical current from one neuron to another, produces wave patterns known as brainwaves. Brainwaves can be categorized into distinct bands based on their frequency \cite{ismail2016}. These include Beta (13-30 Hz), Alpha (8-13 Hz), Theta (4-8 Hz), Delta (0.5-4 Hz), and Gamma (less than 0.5 Hz). Furthermore, some of these are further classified into sub-bands, such as high Alpha, low Alpha, high Beta, low Beta, high Gamma, and low Gamma.

According to the study in ~\cite{Lopez2004}, the Electroencephalogram (EEG) records the oscillations of electric brain potentials obtained through electrodes placed on the human scalp ~\cite{Nunez2006}. These electric potentials directly result from the presence of electric dipoles formed by the postsynaptic potentials generated at the apical dendrites of pyramidal cells in the cortex. The poles of these electric dipoles can be perceived as the source and sink of ionic currents created by excess and deficiency of cations at the soma and apical dendrites, respectively. These ions can move freely through the cerebrospinal fluid and brain tissues, thereby producing ionic currents, which provide the most accurate evidence of the existence of electrical potentials.


For EEG data collection, we utilize the Neurosky MindWave Mobile Headset ~\cite{neuroskyheadset}, a comparatively low-cost, easy-to-use, wearable, portable, and non-invasive brain-computer interface. This headset uses an EEG electrode placed at the FP1 position and the headset's ground electrodes placed on the ear clip. We acquired eight brainwaves namely delta, theta, low alpha, high alpha, low beta, high beta, low gamma, and high gamma which are generated by the Headset at 512 Hz sampling rate. These values have no units and are only meaningful when compared to each other and to themselves.

\section{Related Work}
Numerous neuroimaging studies involving humans have provided compelling evidence of a close connection between the mind and the brain. A study by Haynes et al. ~\cite{Haynes2006} demonstrated progress in reconstructing mental states from noninvasive brain activity measurements, particularly in detecting deception. Lie detection is critical for social interactions, criminal investigations, and national security, yet even experts often struggle with accuracy. Traditional physiological indicators for lie detection include blood pressure, respiration, electrodermal activity, voice stress analysis, and thermal imaging, all of which are influenced by brain activity. In related research, Michael et al. ~\cite{michael2010} showed that religious primes can reduce neurophysiological responses to errors, while Claire et al. ~\cite{claire2017} reported increased gamma brainwave amplitude in three meditation traditions compared to controls.

In yet another study ~\cite{Rocca2014}, researchers proposed a novel approach combining spectral coherence-based connectivity between brain regions as a biometric feature. Tested on 108 subjects in eyes-closed (EC) and eyes-open (EO) resting states, this method showed enhanced distinction compared to power-spectrum measurements. In another study ~\cite{annisa2020personality}, authors used Support Vector Machine (SVM) on EEG-derived features to classify five personality dimensions: neuroticism, extraversion, openness, agreeableness, and conscientiousness. However, the accuracy of this approach was relatively low.

Qin et al. ~\cite{qin_deepforest} introduced an EEG signal recognition method using improved variational mode decomposition (VMD) and deep forest, relying on datasets from epileptic patients without real-time user evaluation. Similarly, Ganaie et al. ~\cite{Ganaie_svm} proposed an improved intuitionistic fuzzy twin support vector machine (IIFTWSVM) to mitigate noise and outliers in EEG signal classification using the same dataset. Hazarike et al. ~\cite{hazarika_svm} employed a different approach with a twin parametric margin SVM based on Universum data (UTPMSVM) for EEG classification. However, none of these studies utilized deep learning models or incorporated real-time user evaluations.

Das Chakladar et al. ~\cite{das_lstm} estimated workload during multitasking mental activities using a hybrid deep-learning framework combining Long Short-Term Memory (LSTM) and Bidirectional Long Short-Term Memory (BiLSTM) models for classification. Hu et al. ~\cite{hu_bilstm} proposed a novel seizure detection method employing BiLSTM for classification using a scalp EEG database. Similarly, Algarni et al. ~\cite{algarni_bilstm} used a BiLSTM-based deep learning approach for emotion recognition from EEG signals.

Our work builds on prior studies by establishing a strong link between human brain activity and individual traits. This relationship enables the development of trait identification techniques, presented to users through our innovative integrated system, allowing them to gain insights into their own traits.
\raggedbottom

\section{Materials and Methods}\label{sec:overview}

\begin{figure}[h]
  \centering
  \includegraphics[width=0.4\textwidth]{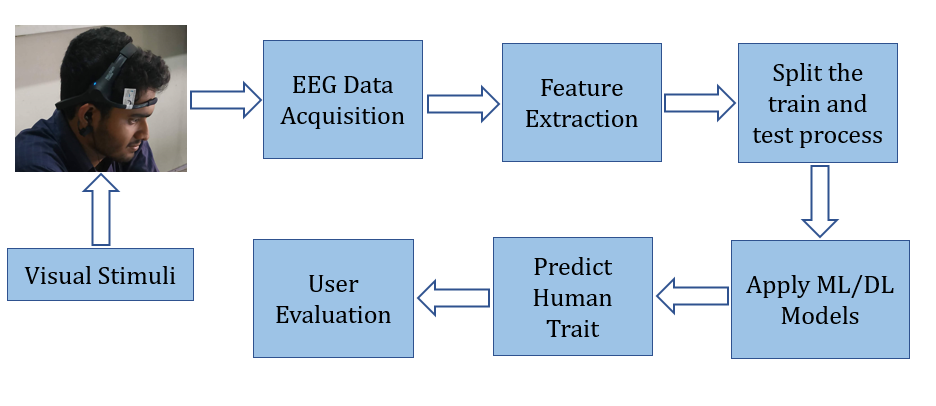}
  \caption{Overview of the methodology}
  \Description{Our Proposed Experimental Model}
  \label{fig:flowchart2}
\end{figure}

We used deep-learning models, LSTM and BiLSTM, to predict human traits from the pre-processed EEG data. These models are suitable for time-series data as they possess the ability to remember both past and recent events, allowing them to predict the target variable accurately. This property of the deep-learning models is particularly useful for predicting human traits based on how EEG signals vary across emotional states. Our methodology is shown in Figure ~\ref{fig:flowchart2}. 

\raggedbottom

\subsection{EEG Data Acquisition Phase}First, we collected EEG signals from subjects using the Neurosky Mindwave Headset. The device generates eight brainwave frequencies, as mentioned earlier: delta (0.5-2.75 Hz), theta (3.5-6.75 Hz), low alpha (7.5-9.25 Hz), high alpha (10-11.75 Hz), low beta (13-16.75 Hz), high beta (18-29.75 Hz), low gamma (31-39.75 Hz), and mid gamma (41-49.75 Hz). During data collection, we presented four emotional videos: happy, sad, neutral, and meditation. The EEG signals corresponding to each emotion were transferred from the headset to a laptop via Bluetooth. As a result, we obtained four different files for each subject, generating a total of 320 samples for 80 subjects.

\subsection{Statistical Analysis}
We conducted a statistical analysis of the data using box plots, shown in Figure 2, which highlight distinct differences in brainwave signals across emotional conditions. The happy condition consistently displays the highest signals, with the delta boxplot showing a median around 25 (in ten thousand) and high variability. In contrast, the sad condition has lower signals, with a delta boxplot median below 20 and reduced variability. Median values decrease across the lower boxplots, indicating minimal brain activity.

\begin{figure*}[ht]
    \centering
    \begin{subfigure}{0.24\textwidth}
        \centering
        \includegraphics[width=\textwidth]{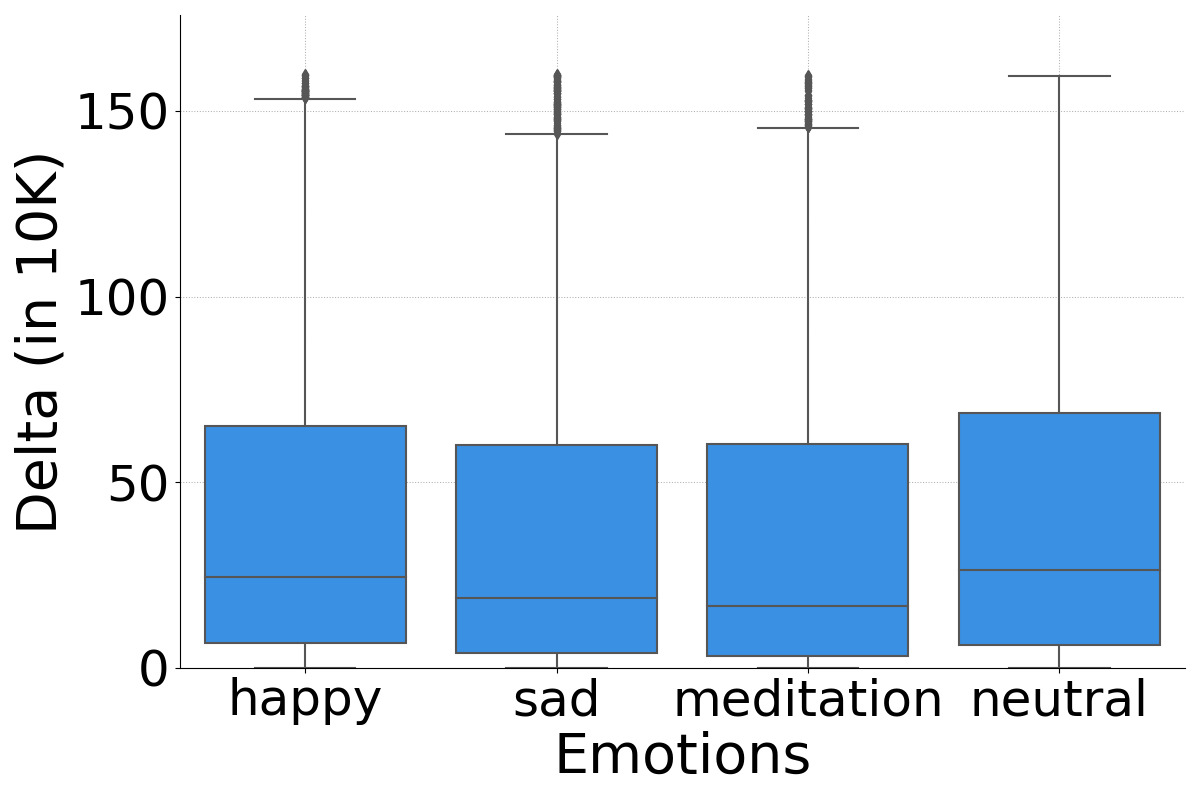}
        \caption{Delta}
        \label{sfig:delta}
    \end{subfigure}
    \hfill
    \begin{subfigure}{0.24\textwidth}
        \centering
        \includegraphics[width=\textwidth]{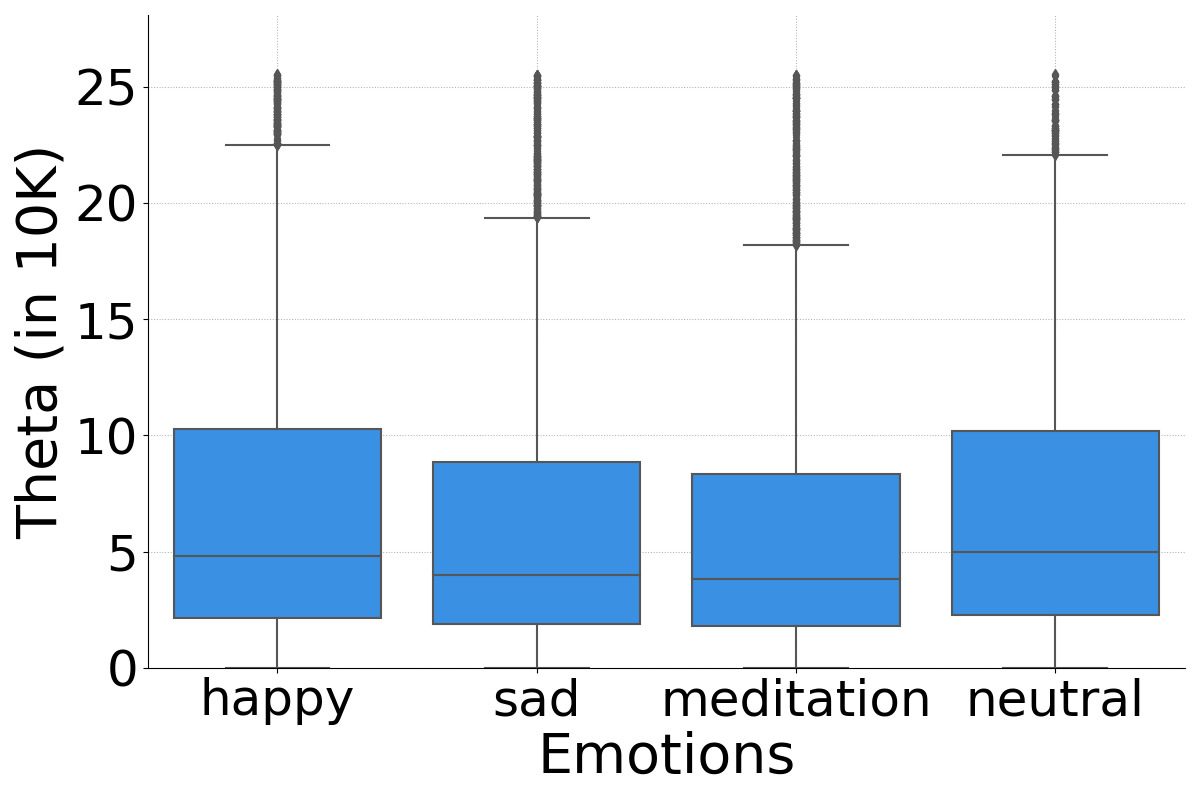}
        \caption{Theta}
        \label{sfig:theta}
    \end{subfigure}
    \hfill
    \begin{subfigure}{0.24\textwidth}
        \centering
        \includegraphics[width=\textwidth]{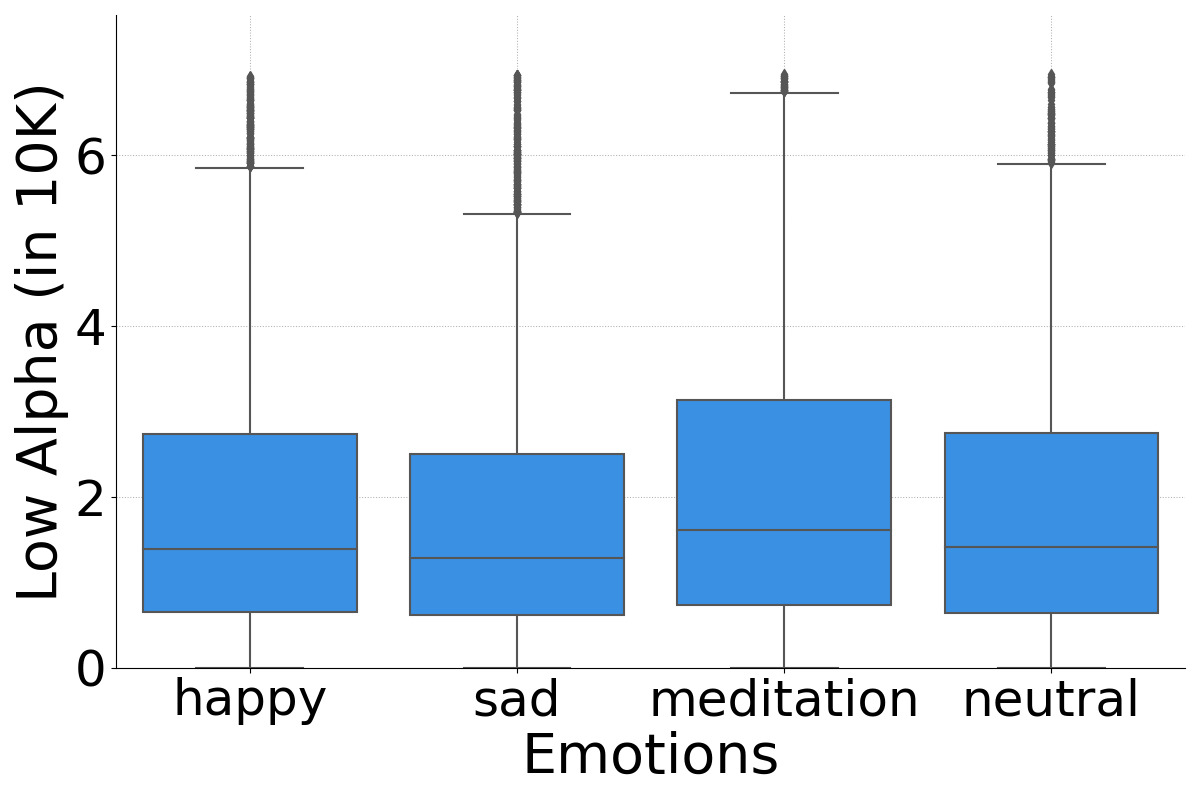}
        \caption{Low Alpha}
        \label{sfig:lowalpha}
    \end{subfigure}
    \hfill
    \begin{subfigure}{0.24\textwidth}
        \centering
        \includegraphics[width=\textwidth]{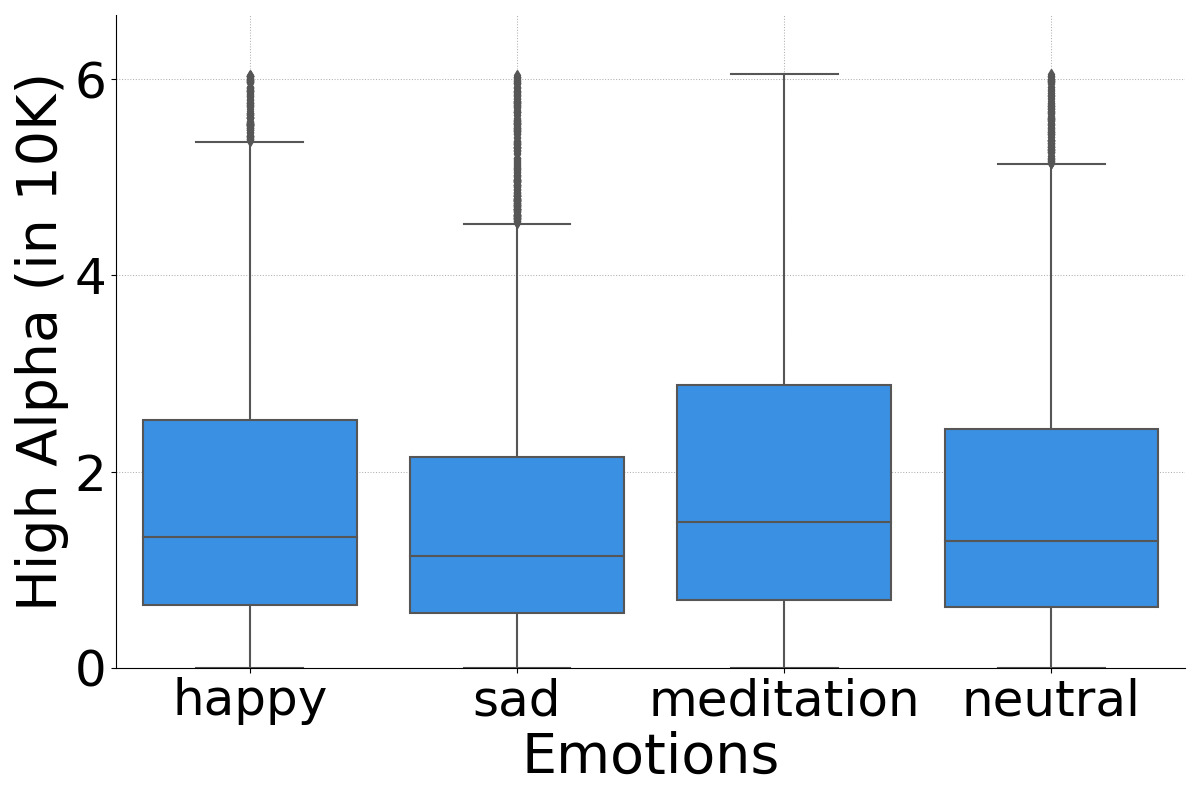}
        \caption{High Alpha}
        \label{sfig:highalpha}
    \end{subfigure}
    \hfill
    \begin{subfigure}{0.24\textwidth}
        \centering
        \includegraphics[width=\textwidth]{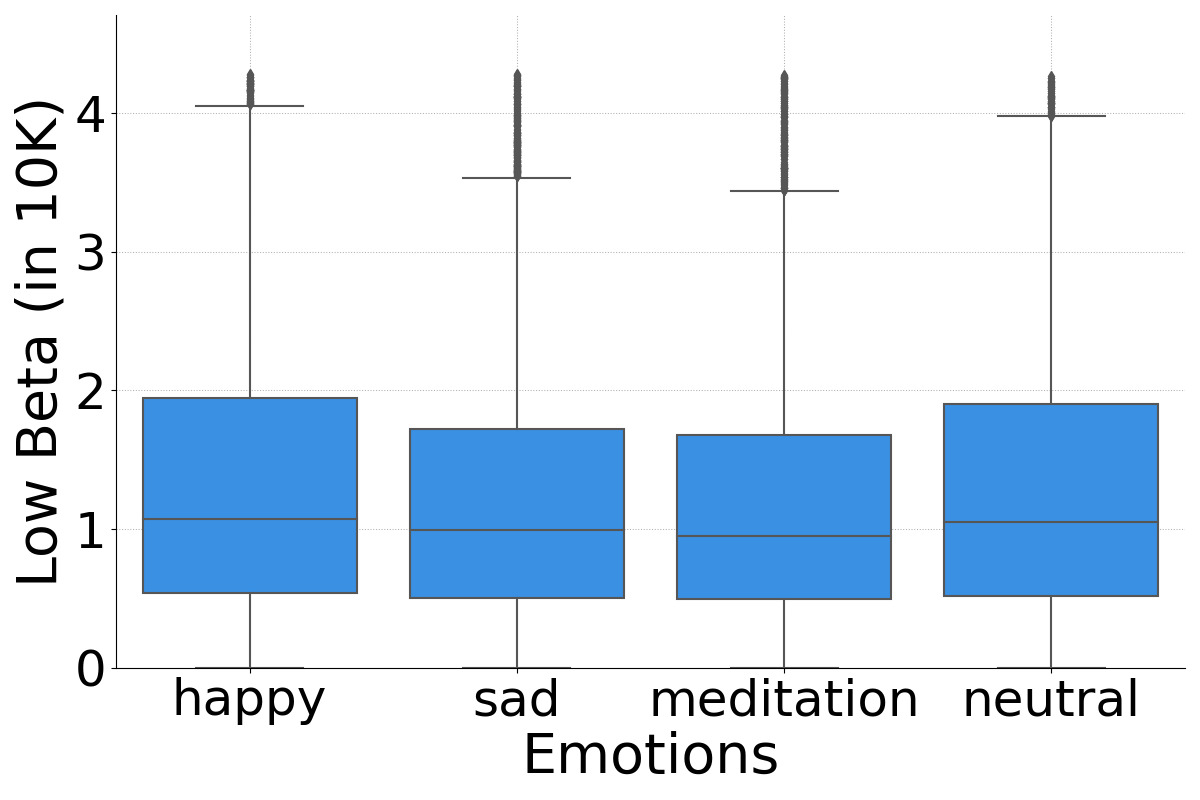}
        \caption{Low Beta}
        \label{sfig:lowbeta}
    \end{subfigure}
    \hfill
    \begin{subfigure}{0.24\textwidth}
        \centering
        \includegraphics[width=\textwidth]{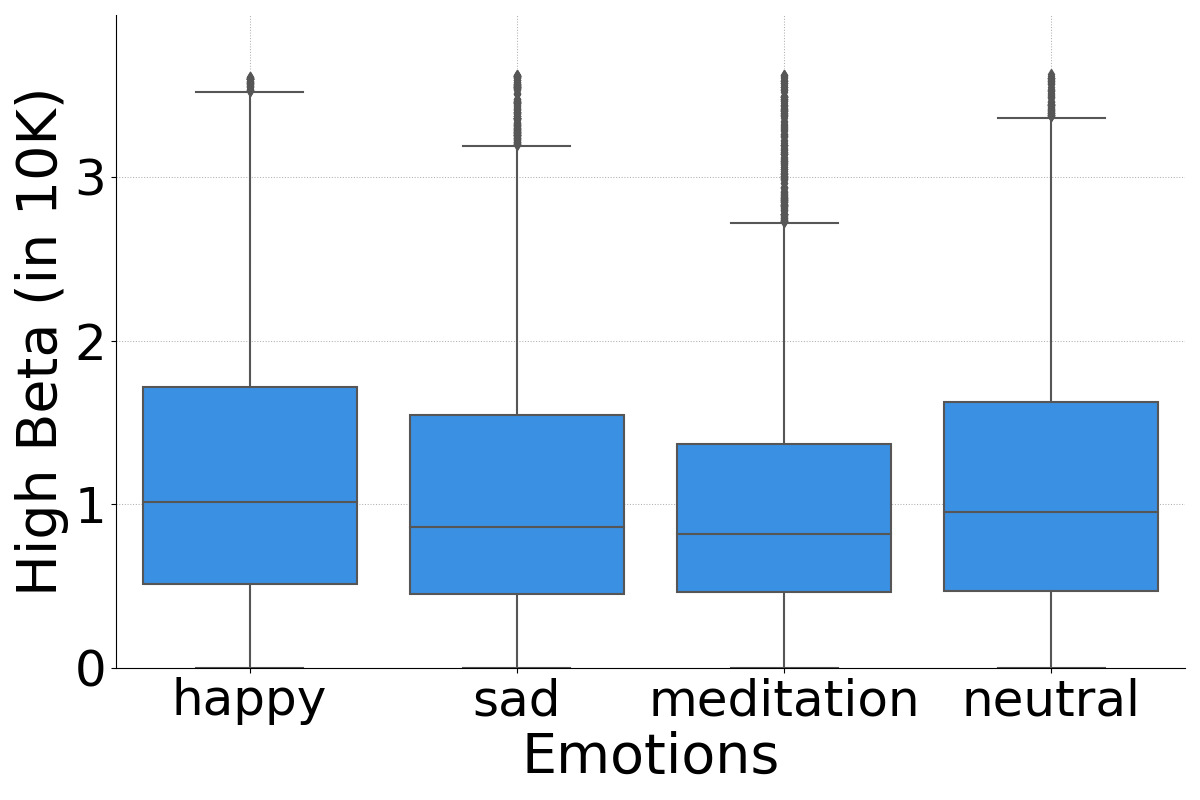}
        \caption{High Beta}
        \label{sfig:highbeta}
    \end{subfigure}
    \hfill
    \begin{subfigure}{0.24\textwidth}
        \centering
        \includegraphics[width=\textwidth]{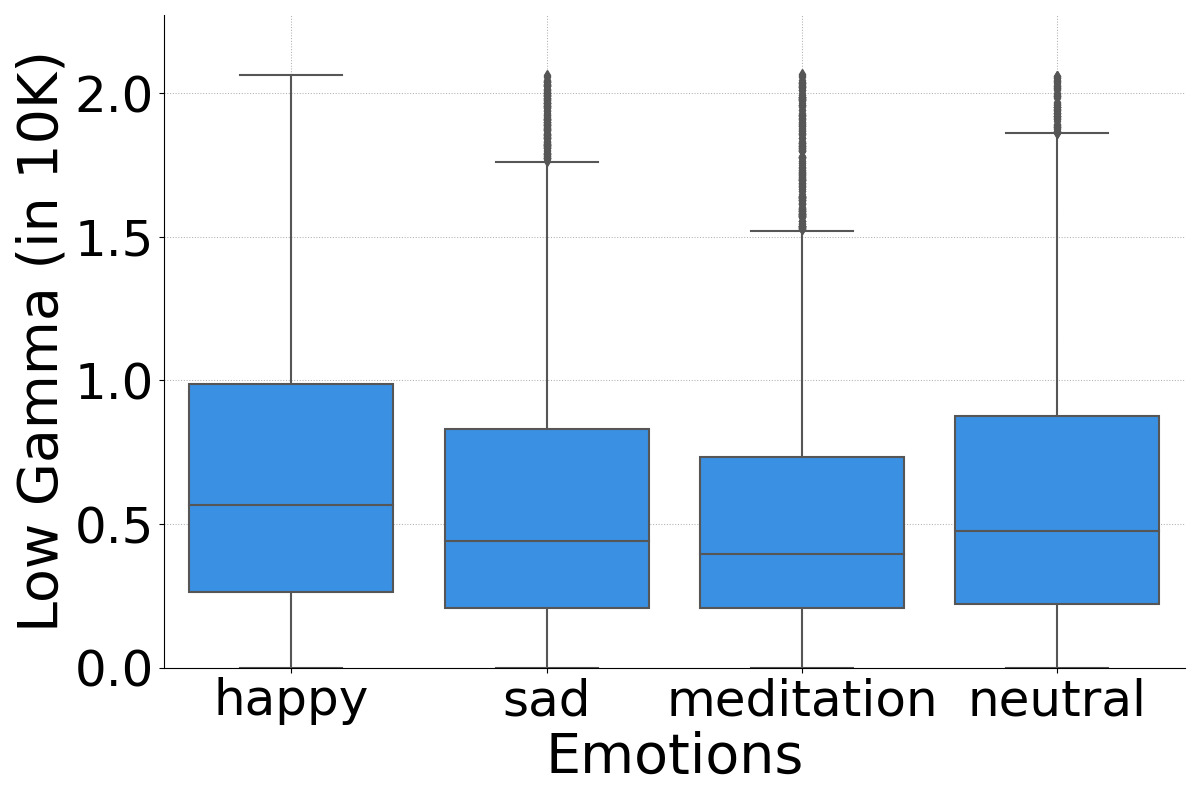}
        \caption{Low Gamma}
        \label{sfig:lowgamma}
    \end{subfigure}
    \hfill
    \begin{subfigure}{0.24\textwidth}
        \centering
        \includegraphics[width=\textwidth]{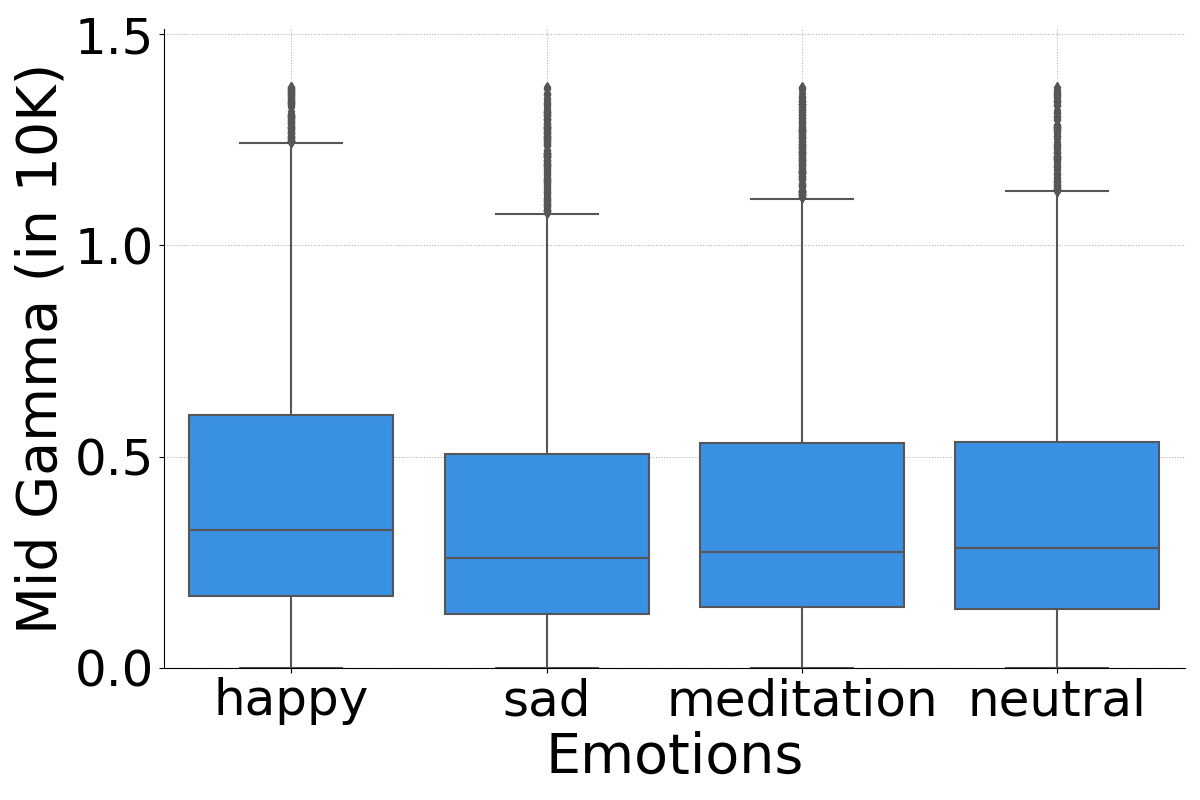}
        \caption{Mid Gamma}
        \label{sfig:midgamma}
    \end{subfigure}
    \hfill
    \caption{Comparison of the relative band powers of different EEG signals for four emotional states: happy, sad, neutral, and meditation.}
    \label{fig:boxplot}
\end{figure*}

The meditation condition exhibits stable signals, with the delta boxplot slightly higher than the sad condition but lower than the happy condition. In the lower graphs, the range narrows, reflecting stable brain signals, particularly in the theta boxplot. The neutral condition initially resembles the happy condition, with a slightly higher median and a broader range, but as the graphs progress, the median decreases and the spread narrows.

\subsection{Feature Extraction}In the feature extraction phase, we calculated two metrics—mean and standard deviation—for each of the eight brainwave bands for every participant. These 2 × 8 (16) data columns were used as features for training. Individually, the features may not exhibit significant traits, but in combination, they uniquely identify human traits, as demonstrated in this paper. Mathematically, each participant’s brainwave data can be represented as a 16-dimensional vector. In this 16-dimensional space, data points from participants cluster into distinct regions, enabling classification of human traits using various classifiers.

\subsection{Train and Test Data Split}Next, We divided the EEG data into training and testing sets using an 80-20 split, with 80\% of the data allocated for training and 20\% for testing. The training set was utilized to train the proposed model, while the testing set assessed its performance on unseen data.

\subsection{Machine Learning Models}
We used Auto-WEKA, a Weka software package, to automatically identify the best classification or regression algorithm for the training dataset. Auto-WEKA was executed on two parallel cores, with a runtime of 20-30 minutes to evaluate a wide range of algorithm combinations.

Brainwave data was collected across four emotional states by displaying distinct videos, resulting in separate brainwave datasets for each state. Survey responses from participants provided ground truth labels for the 14 traits we aimed to train. Each participant's mean and standard deviation of brainwave data were used as features to train individual machine-learning models. In total, we trained 14 models for each of the four emotional states, yielding 56 models.

During the evaluation, trait prediction was based on an aggregated form of the four emotional state models for each trait, selecting the model with the highest training accuracy.

\subsection{Deep Learning Alternatives}
We explored two deep-learning models, Long Short-Term Memory (LSTM) and Bidirectional Long Short-Term Memory (BiLSTM), to compare their performance with the Auto-WEKA package. The EEG dataset was divided into an 80-20 split for training and testing these models.

The LSTM model architecture comprises four layers: a sequential input layer, an LSTM layer with 50 hidden units, a fully connected layer, and a SoftMax layer for output classification. Similarly, the BiLSTM model includes a sequential input layer, a BiLSTM layer with 50 hidden units per direction (forward and backward), effectively totaling 100 units, a fully connected layer, and a SoftMax layer for output classification.




Table \ref{tab:table_lstm} details the deep learning layers, their values, and descriptions used for training the network. The LSTM layer contains 50 units, and the dropout probability is set to 0.2. Similarly, the BiLSTM layer contains 50 units per direction. The model is optimized using the Adam optimizer to minimize the categorical cross-entropy loss function, with a learning rate of 0.001. The maximum number of epochs is set to 50. After training on 80\% of the data, the model was tested on the remaining 20\%. The results of the test are discussed in the next section.


\begin{table}[!ht]
\small
\centering
\caption{Configuration of the Deep Learning Models}
\label{tab:table_lstm}
\resizebox{0.48 \textwidth}{!}{%
\begin{tabular} {|p{0.7in} | p{0.5in} | p{2in} |}
   \hline
\textbf{Deep Learning Layer } & \textbf{ Value} & \textbf{ Description} \\ \hline
LSTM Layer & 50 hidden units & Output Mode: Last \\ \hline
BiLSTM Layer & 50 hidden units & Output Mode: Last \\ \hline
Dropout Layer & 20\% dropout Layer & 20\% units are dropped to prevent over-fitting\\ \hline
Fully Connected Layer & 1 Fully Connected Layer & Transforms the learned features into the final decision space\\ \hline
Classification Layer & SoftMax & Converts raw outputs into a probability distribution over multiple classes \\ \hline
Loss Function & Categorical Cross-Entropy & Computes the error between the predicted probabilities and the true labels\\ \hline
ADAM & - & Adaptive Moment Estimation-Optimization Algorithm\\ \hline
No. of Epochs & 50 & Models go through the whole dataset 50 times\\ \hline
Batch Size & 32 & Total samples are divided batches, each batch having 32 samples\\ \hline
Initial Learning Rate & 0.001 & Models' weights are updated by 0.1\% of the gradient\\  \hline 
\end{tabular}
}
\end{table}


\raggedbottom

\subsection{Trait Identification Process}
After training, trait identification is performed by collecting brainwave data from a random individual. This data represents a point in the 16-dimensional space, enabling trait identification. During testing, we evaluated the model using brainwave data from 20 additional individuals to determine their traits. The proposed classification technique identifies human traits based on the emotional states used during training.

\subsection{User Evaluation}
To evaluate the performance of our proposed technique, we assessed our automated system, which collects user brainwave data while displaying videos and predicts their traits via a Java application. Users rate each prediction as 1 for correct or 0 for incorrect. All computations are performed within the Java application, which acquires the user's brainwave dataset to identify traits.

The experimental setup mirrors the training phase, with the addition of immediately displaying trait prediction results after data collection. Predictions are generated by loading trained machine-learning models for each human trait and processing the user's brainwave data. Users evaluate the predictions, assigning a score of 1 for correct or 0 for incorrect predictions. A total of 14 traits and behaviors are predicted, and these evaluations are used to calculate test accuracy.

\section{Experimental Findings}
This section presents the experimental findings. Our results demonstrate that the proposed technique achieves reasonably high training accuracy for specific trait-emotion pairs across multiple machine-learning algorithms.


As detailed in the preceding section, we trained 70 machine-learning models, with their training accuracies summarized in Table \ref{table_merged}. The table shows exceptionally high training accuracy for identifying religious believers, individuals engaging in regular exercise, and families with a history of heart disease across all emotional states. Additionally, we achieve strong accuracy in identifying smokers in happy and sad states, practitioners in all states except neutral, stroke cases in meditation and neutral states, and diabetes cases in all states except meditation and neutral. Conversely, the models show relatively poor accuracy in identifying fast food intake across all emotional states.



\input{table-merged}

We also trained the LSTM and BiLSTM models to compare their performance with the Auto-WEKA models. The training accuracies of these deep-learning models are also shown in Table \ref{table_merged}. In most cases, the Auto-WEKA models outperformed the deep-learning models. Consequently, we selected the Auto-WEKA models for user evaluation, which is discussed in the next section.

Table \ref{tab:qualitative} shows the comparison of our proposed method with recent studies. Most of the recent studies use previously collected open-source data. We use our own collected data after a brief data-collection process. We also explore both machine-learning and deep-learning models for our experiment. Other studies explore one of them. We also use a Java application for real-time user evaluation which is missing in other recent studies. 
\raggedbottom

\begin{table}[h]
\centering
\caption{Comparison of Our Proposed Approach With Other Existing Research Studies}
\label{tab:qualitative}
\small
\begin{tabular}{c|c|c|c|c}
\hline
Name & \begin{tabular}[c]{@{}c@{}}Data\\ Collection\end{tabular} & \begin{tabular}[c]{@{}c@{}}Machine\\ Learning\end{tabular} & \begin{tabular}[c]{@{}c@{}}Deep \\ Learning\end{tabular} & \begin{tabular}[c]{@{}c@{}}Real-time \\ User\\ Evaluation\end{tabular} \\ \hline
Qin et al., \cite{qin_deepforest} &  & \ding{51} &  &  \\ 
Ganaie et al., \cite{Ganaie_svm} &  & \ding{51} &  &  \\
Hazarika et al., \cite{hazarika_svm} &  & \ding{51} &  &  \\
Chakladar et al., \cite{das_lstm} &  &  & \ding{51} &  \\
Hu et al., \cite{hu_bilstm} &  &  & \ding{51} &  \\
Algarni et al., \cite{algarni_bilstm} &  &  & \ding{51} &  \\
\textbf{\begin{tabular}[c]{@{}l@{}}Our Proposed \\ Approach\end{tabular}} & \ding{51} & \ding{51} & \ding{51} & \ding{51} \\ \hline
\end{tabular}
\end{table}

\section{Conclusion and Future Work}
A real-time system for identifying human traits has potential applications in areas such as medical diagnosis, security, and healthcare. Traits, often modified or hidden, can be accurately identified using this system. However, it has limitations, such as the need for the EEG headset to stay within Bluetooth range, which presents opportunities for improvement.

Data size and diversity are critical for prediction models. In the future, we aim to expand data collection, ensure diversity, and generate synthetic data to improve robustness. The system could also be applied to disease detection via brainwave analysis, reducing diagnostic costs.

While we analyzed 14 traits, future work will explore additional traits correlated with brain activity. To reduce noise in EEG signals, we plan to use more reliable mechanisms with additional electrodes. We also aim to replace traditional machine-learning algorithms with deep neural networks, improving classification accuracy through iterative enhancements.



\bibliography{sample-base}

%

\end{document}

%% file: table-merged.tex
\begin{table*}[!ht]
    \centering
    \caption{Machine Learning Models Accuracy using Auto-WEKA package and Deep Learning}
    \resizebox{0.49 \textwidth}{!}{%
    \begin{tabular}{|l|c|c|c|c|}
        \hline
        \multicolumn{1}{|l|}{\multirow{2}{*}{\textbf{\begin{tabular}[c]{@{}l@{}}Predicted Trait\\  - Emotion\end{tabular}}}} & \multicolumn{2}{c|}{\textbf{\begin{tabular}[c]{@{}c@{}}Auto-Weka \\ Accuracy (\%)\end{tabular}}} & \multicolumn{2}{c|}{\textbf{\begin{tabular}[c]{@{}c@{}}Deep Learning \\ Models Accuracy (\%)\end{tabular}}} \\ \cline{2-5} 
        \multicolumn{1}{|l|}{}                                                                                               & \multicolumn{1}{c|}{\textbf{Classifier}}         & \multicolumn{1}{c|}{\textbf{Accuracy}}        & \multicolumn{1}{c|}{\textbf{LSTM}}                  & \multicolumn{1}{c|}{\textbf{BiLSTM}}                  \\ \hline
        Smoker - Happy            & LWL                                                                     & 70.67 & 69.23 & 61.54 \\ \hline
Smoker - Meditation       & Random Tree                                                             & 59.21 & 46.15 & 38.46 \\ \hline
Smoker - Neutral          & J48                                                                     & 65.28 & 50.00 & 66.67 \\ \hline
Smoker - Sad              & JRip                                                                    & 71.05 & 42.86 & 35.71 \\ \hline
Alcoholic - Happy         & Naïve Bayes                                                             & 62.67 & 61.54 & 53.85 \\ \hline
Alcoholic - Meditation    & \begin{tabular}[c]{@{}l@{}}Naïve Bayes\\ Multinomial\end{tabular}       & 61.84 & 53.85 & 46.15 \\ \hline
Alcoholic - Neutral       & Simple Logistic                                                         & 65.28 & 66.67 & 58.33 \\ \hline
Alcoholic - Sad           & AdaBoost M1                                                             & 67.11 & 42.86 & 35.71 \\ \hline
Believer - Happy          & Voted Perceptron                                                        & 89.33 & 92.31 & 69.23 \\ \hline
Believer - Meditation     & Voted Perceptron                                                        & 89.47 & 84.62 & 76.92 \\ \hline
Believer - Neutral        & SMO                                                                     & 90.28 & 75.00 & 91.67 \\ \hline
Believer - Sad            & OneR                                                                    & 93.42 & 78.57 & 71.43 \\ \hline
Practitioner - Happy      & AdaBoost M1                                                             & 76.32 & 53.85 & 30.77 \\ \hline
Practitioner - Meditation & J48                                                                     & 77.63 & 53.85 & 46.15 \\ \hline
Practitioner - Neutral    & J48                                                                     & 44.44 & 41.67 & 41.67 \\ \hline
Practitioner - Sad        & Bagging                                                                 & 76.35 & 42.86 & 35.71 \\ \hline
Exercise - Happy          & Bagging                                                                 & 82.67 & 92.31 & 92.31 \\ \hline
Exercise - Meditation     & PART                                                                    & 81.58 & 84.62 & 92.31 \\ \hline
Exercise - Neutral        & OneR                                                                    & 90.28 & 75.00 & 83.33 \\ \hline
Exercise - Sad            & Vote                                                                    & 82.89 & 64.29 & 78.57 \\ \hline
Fast food - Happy         & Simple Logistic                                                         & 46.67 & 46.15 & 46.15 \\ \hline
Fast food - Meditation    & Decision Stump                                                          & 51.32 & 69.23 & 38.46 \\ \hline
Fast food - Neutral       & Bagging                                                                 & 47.22 & 50.00 & 41.67 \\ \hline
Fast food - Sad           & LWL                                                                     & 63.16 & 57.14 & 35.71 \\ \hline
Fat - Happy               & Bagging                                                                 & 50.67 & 46.15 & 46.15 \\ \hline
Fat - Meditation          & Decision Stump                                                          & 65.79 & 61.54 & 46.15 \\ \hline
Fat - Neutral             & \begin{tabular}[c]{@{}l@{}}Attribute Selected\\ Classifier\end{tabular} & 55.56 & 50.00 & 41.67 \\ \hline
Fat - Sad                 & Multilayer Perceptron                                                   & 59.21 & 71.43 & 78.57 \\ \hline
        \end{tabular}
        }
        \resizebox{0.49 \textwidth}{!}{%
        \begin{tabular}{|l|c|c|c|c|}
        \hline
        \multicolumn{1}{|l|}{\multirow{2}{*}{\textbf{\begin{tabular}[c]{@{}l@{}}Predicted Trait\\  - Emotion\end{tabular}}}} & \multicolumn{2}{c|}{\textbf{\begin{tabular}[c]{@{}c@{}}Auto-Weka \\ Accuracy (\%)\end{tabular}}} & \multicolumn{2}{c|}{\textbf{\begin{tabular}[c]{@{}c@{}}Deep Learning \\ Models Accuracy (\%)\end{tabular}}} \\ \cline{2-5} 
        \multicolumn{1}{|l|}{}                                                                                               & \multicolumn{1}{c|}{\textbf{Classifier}}         & \multicolumn{1}{c|}{\textbf{Accuracy}}        & \multicolumn{1}{c|}{\textbf{LSTM}}                  & \multicolumn{1}{c|}{\textbf{BiLSTM}}                  \\ \hline
        Sugar - Happy                 & SMO                                                             & 62.67 & 61.54  & 61.54 \\ \hline
Sugar - Meditation            & SMO                                                             & 63.16 & 69.23  & 53.85 \\ \hline
Sugar - Neutral               & \begin{tabular}[c]{@{}l@{}}Random\\ Sub Space\end{tabular}      & 63.89 & 66.67  & 75.00 \\ \hline
Sugar - Sad                   & SMO                                                             & 65.79 & 50.00  & 57.14 \\ \hline
Vegetable - Happy             & Logistic                                                        & 70.67 & 76.92  & 84.62 \\ \hline
Vegetable - Meditation        & Logistic                                                        & 72.37 & 76.92  & 69.23 \\ \hline
Vegetable - Neutral           & \begin{tabular}[c]{@{}l@{}}Multilayer\\ Perceptron\end{tabular} & 79.17 & 58.33  & 83.33 \\ \hline
Vegetable - Sad               & J48                                                             & 76.32 & 71.43  & 57.14 \\ \hline
Sleep time - Happy            & BayesNet                                                        & 54.67 & 69.23  & 53.85 \\ \hline
Sleep time - Meditation       & Decision Stump                                                  & 53.95 & 53.85  & 38.46 \\ \hline
Sleep time - Neutral          & Decision Stump                                                  & 52.78 & 41.67  & 66.67 \\ \hline
Sleep time - Sad              & REPTree                                                         & 53.95 & 64.29  & 35.71 \\ \hline
Sleeping Problem - Happy      & REPTree                                                         & 58.67 & 61.54  & 61.54 \\ \hline
Sleeping Problem - Meditation & JRip                                                            & 46.05 & 46.15  & 61.54 \\ \hline
Sleeping Problem - Neutral    & Bagging                                                         & 44.44 & 41.67  & 25.00 \\ \hline
Sleeping Problem - Sad        & Decision Stump                                                  & 53.95 & 57.14  & 42.86 \\ \hline
Heart disease- Happy          & PART                                                            & 76.00 & 61.54  & 76.92 \\ \hline
Heart disease- Meditation     & J48                                                             & 81.58 & 76.92  & 92.31 \\ \hline
Heart disease - Neutral       & PART                                                            & 75.00 & 83.33  & 66.67 \\ \hline
Heart disease - Sad           & Decision Stump                                                  & 77.63 & 64.29  & 92.86 \\ \hline
Diabetes - Happy              & Bagging                                                         & 60.00 & 69.23  & 76.92 \\ \hline
Diabetes - Meditation         & Decision Table                                                  & 59.21 & 69.23  & 69.23 \\ \hline
Diabetes - Neutral            & LWL                                                             & 94.44 & 66.67  & 66.67 \\ \hline
Diabetes - Sad                & Bagging                                                         & 65.79 & 57.14  & 50.00 \\ \hline
Stroke - Happy                & Decision Stump                                                  & 57.33 & 92.31  & 76.92 \\ \hline
Stroke - Meditation           & IBk                                                             & 92.11 & 84.62  & 92.31 \\ \hline
Stroke - Neutral              & LWL                                                             & 69.44 & 83.33  & 83.33 \\ \hline
Stroke - Sad                  & Decision Stump                                                  & 56.58 & 100.00 & 85.71 \\ \hline
        \end{tabular}
        }
    \label{table_merged}
\end{table*}